\documentclass{article}

\usepackage{arxiv}

\usepackage{graphicx}
\usepackage{algorithm}
\usepackage{algpseudocode}
\usepackage{graphicx}   % 基本的插图
\usepackage{subcaption} % 子图支持
\usepackage{hyperref}
\usepackage{booktabs}
\usepackage{multirow}
\usepackage{wrapfig}
\usepackage{soul}
\usepackage{bm}
\usepackage{pifont}
\usepackage{amsmath}
\usepackage{amsfonts}  % or \usepackage{amssymb}

\newcommand{\cmark}{\checkmark}  % ✓
\newcommand{\xmark}{\ding{55}}   % ✗
\usepackage{url}
\usepackage{enumitem}
\usepackage{adjustbox}
\usepackage{makecell}
\usepackage{threeparttable}

\begin{document}

\title{
ReasonLight: A Multimodal Foundation Model-Enhanced
Reinforcement Learning Framework for
Zero-Shot Traffic Signal Control
}

\author{
Aoyu~Pang$^{1}$,
Maonan~Wang$^{2,3}$,
Yuejiao~Xie$^{1}$,
Chung~Shue~Chen$^{4}$,
Zhiwei~Yang$^{1}$,
and Man-On~Pun$^{1}$\\[0.6em]
\small
$^{1}$School of Science and Engineering, The Chinese University of Hong Kong, Shenzhen, China\\
\small
$^{2}$Department of Mechanical and Automation Engineering, The Chinese University of Hong Kong, Hong Kong\\
\small
$^{3}$Shanghai AI Laboratory, Shanghai, China\\
\small
$^{4}$Nokia Bell Labs, Paris-Saclay, France
}

\maketitle

\begin{abstract}
Reinforcement learning (RL) has shown promise in traffic signal control (TSC). However, its reliance on predefined states limits responsiveness to observable open-world events that are absent from training data. IoT-enabled intersections provide heterogeneous observations from roadside sensors and cameras, creating opportunities to improve RL adaptability to such event. To this end, we propose ReasonLight—a multimodal foundation model-enhanced RL framework for zero-shot TSC. ReasonLight integrates three sources of information: structured traffic measurements, multi-view camera observations, and candidate phase decisions from a pre-trained RL controller. Given an RL-proposed phase, ReasonLight extracts visual semantics from multi-view images and aligns them with compact sensor-derived scene descriptions. This alignment enables a semantic-guided refinement module to either preserve or adjust the proposed action according to traffic rules and event semantics. To ensure operational reliability, refined actions are constrained by the set of available phases. Any invalid decisions rejected by falling back to the original RL action. We evaluate ReasonLight on two types of rare events not seen during RL training: emergency vehicle priority and temporary traffic regulation. Experimental results show that ReasonLight achieves zero-shot adaptation without retraining. It reduces emergency vehicle waiting time by up to 88.7\% compared the RL-only backbone while preserving comparable routine traffic performance.
\end{abstract}

% %%%%%%%%%%%%%
% Introduction
% %%%%%%%%%%%%%
\section{Introduction}

\begin{figure}[!t]
    \centering
    \includegraphics[width=0.6\linewidth]{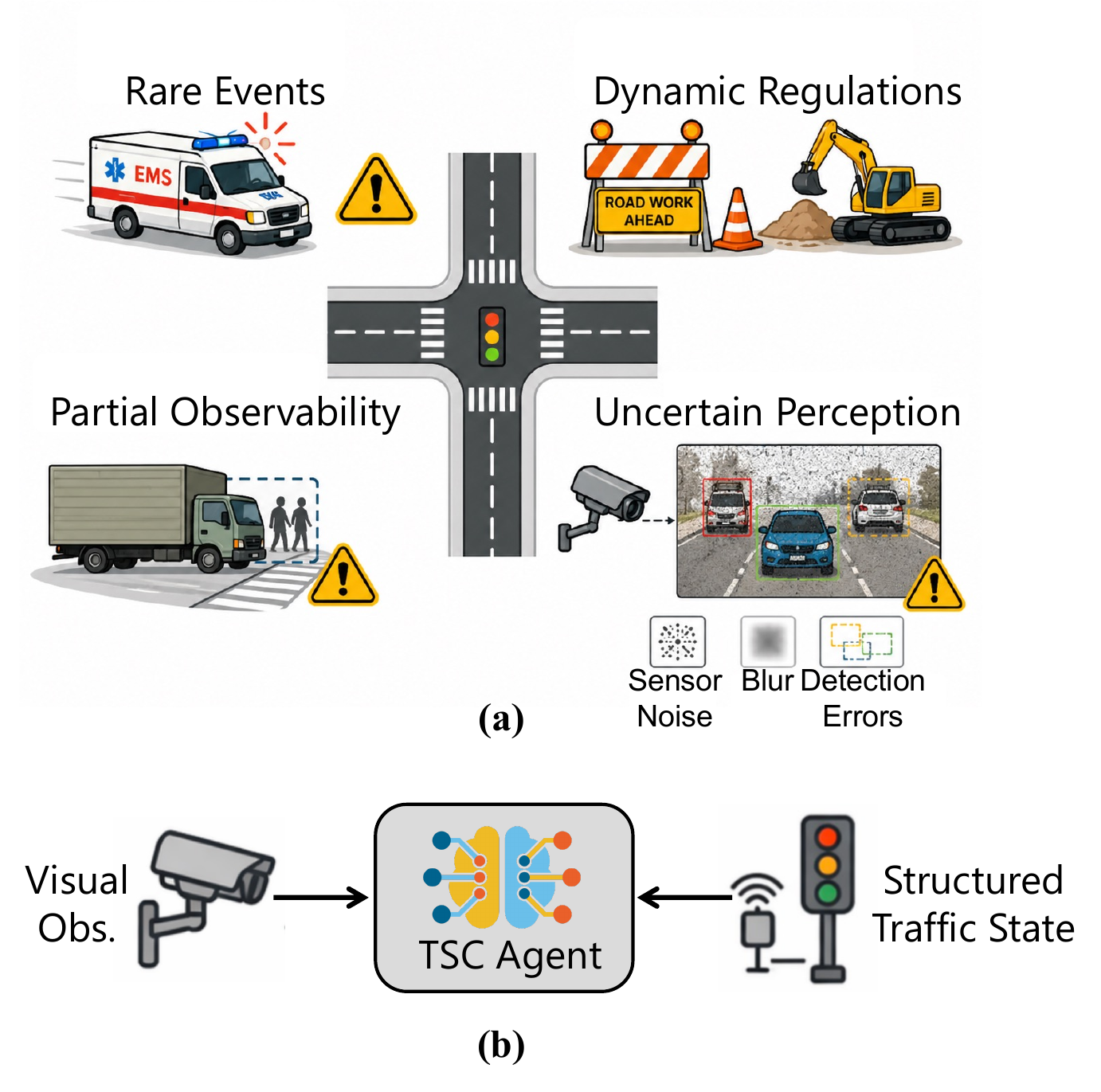}
    \caption{Open-world traffic scenarios and multimodal IoT sensing for zero-shot TSC. (a) Complex and long-tail situations such as emergency vehicle priority and temporary traffic regulations require visual semantics that are not captured by predefined structured states alone. (b) ReasonLight combines structured traffic measurements from roadside sensors with multi-view camera observations, grounding rare-event reasoning in both quantitative traffic states and visual context.}
    \label{fig:scenario}
\end{figure}

Traffic signal control (TSC) is a critical component of modern intelligent transportation systems for mitigating urban congestion. Traditional TSC approaches primarily relied on rule-based or adaptive systems that adjust signal timings based on fixed control logic or historical statistics \cite{lowrie1990scats, koonce2008traffic}. However, these methods often struggle to cope with highly dynamic traffic fluctuations. To address this, reinforcement learning (RL) has emerged as a prominent data-driven approach, demonstrating strong performance by interacting with environments to optimize long-term traffic metrics \cite{wei2019survey, zhai2025hgat}. Despite this success, existing RL-based TSC frameworks fundamentally rely on predefined, low-dimensional structured state representations, such as queue lengths and vehicle counts \cite{noaeen2022reinforcement, zhao2024survey, 10472584, pangreinforcement, wang2024ccda, wang2024unitsa}. This inherent representational bottleneck limits the ability of RL agents to perceive and respond to open-world uncertainties. In reality, intersections frequently encounter complex and long-tail situations, including emergency vehicles, temporary traffic regulations, occlusions, and noisy observations \cite{dulac2021challenges, chen2024efficient}, as illustrated in Fig.~\ref{fig:scenario}(a). Since these events are unpredictable and often underrepresented during training, conventional RL agents can exhibit brittle zero-shot generalization, failing to adapt to previously unseen traffic conditions without costly and unsafe retraining.

To bridge the gap between closed-world training and open-world deployment, IoT-enabled intersections provide an increasingly rich sensing substrate: roadside units and loop detectors offer structured measurements such as flow, occupancy, and signal status, while intersection cameras capture visual events that are difficult to encode as predefined numerical states. Multimodal foundation models (FMs)~\cite{li2024multimodal}, particularly vision-language models (VLMs)~\cite{bai2025qwen3}, provide a promising tool for interpreting such visual observations \cite{pang2024illm, cui2025trafficllm, lai2025llmlight, yuan2025collmlight, zou2025traffic}. By processing camera views, VLMs can recognize semantic traffic events such as approaching emergency vehicles, temporary lane closures, and abnormal road conditions. However, vision-dominant or language-dominant controllers remain insufficient for real-time TSC because they often lack reliable access to fine-grained quantitative traffic dynamics, including lane-level flow rates, queue lengths, and occupancy patterns~\cite{meng2025sl}. Moreover, language-based decision modules may produce unstable or invalid phase decisions when they are not grounded in structured traffic states and admissible signal constraints~\cite{9653682}.

These observations suggest that open-world TSC should not rely on either structured sensing or visual reasoning alone. Instead, a practical IoT traffic controller should combine their complementary strengths. As depicted in Fig.~\ref{fig:scenario}(b), structured infrastructure measurements provide stable quantitative grounding for routine control, while camera-based visual semantics expose rare events that are invisible to standard RL state vectors. In this work, we therefore study a policy-agnostic refinement paradigm: an existing RL controller remains responsible for routine optimization, and a multimodal FM is introduced only as a constrained semantic evaluator that can refine the RL-proposed phase when visual and sensor-derived context indicates an unseen event.

Motivated by this synergy, we propose \textbf{ReasonLight}, a multimodal FM-enhanced RL framework for zero-shot TSC in IoT-enabled intersections. ReasonLight is designed to be compatible with existing RL-based TSC algorithms: it only requires the candidate phase output by an RL controller and does not modify the original policy architecture or reward function. During inference, multi-view camera observations are analyzed by a VLM to extract visual semantics, while structured sensor measurements are converted into a compact textual scene description. A semantic-guided action refinement module then jointly considers the visual semantics, sensor-derived scene description, and RL-proposed action under basic traffic rules and available phase constraints. The main contributions are summarized as follows:
\begin{itemize}[leftmargin=*]
    \item We propose ReasonLight, an IoT-oriented multimodal TSC framework that fuses camera-based visual semantics with structured traffic measurements from roadside infrastructure. This design enables the controller to recognize visually observable long-tail events while preserving quantitative grounding from traffic sensors.
    \item We develop a policy-agnostic semantic-guided action refinement module that can be attached to existing RL controllers without retraining or reward modification. The module evaluates the RL-proposed phase using multimodal context and only outputs actions from the available traffic phase set, with fallback to the RL decision for invalid outputs.
    \item We evaluate ReasonLight on two fully unseen open-world scenarios using only basic traffic-rule prompts during inference. Experiments show that combining visual semantics with sensor-derived scene descriptions provides a more balanced and reliable decision basis than either RL-only or vision-dominant control.
\end{itemize}

The remainder of this paper is organized as follows. Section~\ref{sec:related_work} reviews related work. Section~\ref{sec:preliminaries} introduces the preliminaries and open-world traffic scenarios. Section~\ref{sec:method} details the proposed ReasonLight framework. Section~\ref{sec:experiment} reports the experimental evaluation, and Section~\ref{sec:conclusion} concludes the paper.

% %%%%%%%%%%%%%
% Related Work
% %%%%%%%%%%%%%
\section{Related Work} \label{sec:related_work}

\noindent\textbf{RL-Based Traffic Signal Control.} 
Recent research on TSC has increasingly focused on RL to enable adaptive and real-time decision making. Traditional rule-based methods~\cite{lowrie1990scats, koonce2008traffic} have been widely deployed in urban areas but often fail to adapt to the stochastic and dynamic nature of real-world traffic, resulting in suboptimal performance. To address this limitation, RL has been introduced into TSC to enable adaptive and real-time decision-making~\cite{pang2024scalable, oroojlooy2020attendlight, wang2024unitsa, 11012664}. RL-based methods typically deploy trainable agents at intersections that dynamically adjust signals based on real-time traffic states and have demonstrated promising results in simulation studies. However, deploying these methods in real-world scenarios remains challenging due to issues such as long-tail events~\cite{su2022emvlight} and noisy or missing observations~\cite{feng2024difflight}, which are often underrepresented during training. To bridge the gap between controlled simulations and real-world complexity, this work proposes integrating FMs into RL-based TSC frameworks, leveraging their generalization and representation capabilities to enhance robustness and adaptability.

\begin{table}[!t]
\centering
\caption{Comparison with representative language- and vision-based TSC methods.}
\label{tab:method_comparison}
\begin{tabular}{@{}lccccc@{}}
\toprule
\multirow{2}{*}{Method} 
& \multicolumn{2}{c}{Input modalities} 
& \multicolumn{3}{c}{RL usage} \\
\cmidrule(lr){2-3} \cmidrule(lr){4-6}
& \shortstack{Visual\\input} 
& \shortstack{Structured\\state} 
& \shortstack{Uses\\RL} 
& \shortstack{Role of\\RL} 
& \shortstack{Reusable\\backbone} \\
\midrule
\multicolumn{6}{@{}l}{\textit{LLM-based methods}} \\
LA-Light~\cite{wang2024llm} 
& \xmark & \cmark & \cmark & Tool & \cmark \\
iLLM-TSC~\cite{pang2024illm} 
& \xmark & \cmark & \cmark & Backbone & \cmark \\
LLMLight~\cite{lai2025llmlight} 
& \xmark & \cmark & \xmark & -- & -- \\
Traffic-R1~\cite{zou2025traffic} 
& \xmark & \cmark & \xmark & -- & -- \\
\midrule
\multicolumn{6}{@{}l}{\textit{VLM-based methods}} \\
VLMLight~\cite{wangvlmlight} 
& \cmark & \xmark & \cmark & Coupled & \xmark \\
\midrule
\multicolumn{6}{@{}l}{\textit{Ours}} \\
ReasonLight 
& \cmark & \cmark & \cmark & Backbone & \cmark \\
\bottomrule
\end{tabular}
\end{table}

\noindent\textbf{LLM-Based Traffic Signal Control.}
Developing TSC systems with strong adaptability remains a longstanding challenge. Recent studies have explored the use of LLMs to enhance high-level reasoning and decision making in TSC tasks~\cite{wang2024llm, pang2024illm, lai2025llmlight, zou2025traffic}. LA-Light~\cite{wang2024llm} formulates the LLM as an AI agent that can reason over complex traffic contexts and invoke external traffic-control tools, including RL-based controllers, to support human-mimetic signal decisions. iLLM-TSC~\cite{pang2024illm} further integrates LLMs with RL by first allowing an RL agent to generate a signal decision and then using the LLM to evaluate and revise unreasonable actions under degraded observations or rare events. More recent methods, such as LLMLight~\cite{lai2025llmlight} and Traffic-R1~\cite{zou2025traffic}, investigate LLMs as traffic signal agents or reinforced reasoning models for improving generalization and decision interpretability. Despite these advances, most LLM-based TSC methods rely primarily on structured or textual traffic descriptions and do not directly exploit multi-view visual observations from camera-equipped IoT intersections. As a result, visually grounded events such as emergency vehicles, roadwork signs, temporary barriers, or lane closures must be manually encoded or indirectly inferred from non-visual states.

\noindent\textbf{Multimodal Foundation Models for Traffic Signal Control.}
Multimodal FMs, including VLMs, have demonstrated strong perception and cross-modal reasoning capabilities across various domains~\cite{zhang2024improve, xu2024vlm, yin2024survey, zhao2025vlm, 11261671, 11418645}. Their ability to interpret camera observations makes them attractive for open-world TSC, where rare events are often visually observable before they are reflected in structured traffic states. Recent VLM-based approaches, such as VLMLight~\cite{wangvlmlight}, show that visual scene understanding can improve safety-critical traffic control. However, vision-dominant control can be unreliable for fine-grained traffic optimization, as VLMs may misestimate vehicle counts, queues, and lane occupancy in complex scenes. As summarized in Table~\ref{tab:method_comparison}, existing LLM-based methods mainly rely on structured or textual inputs, while VLM-based methods introduce visual perception but are often tied to specific control pipelines. In contrast, ReasonLight grounds visual semantics in structured traffic states and treats RL as a reusable backbone for multimodal action refinement.

\begin{figure}[!t]
    \centering
    \includegraphics[width=0.75\linewidth]{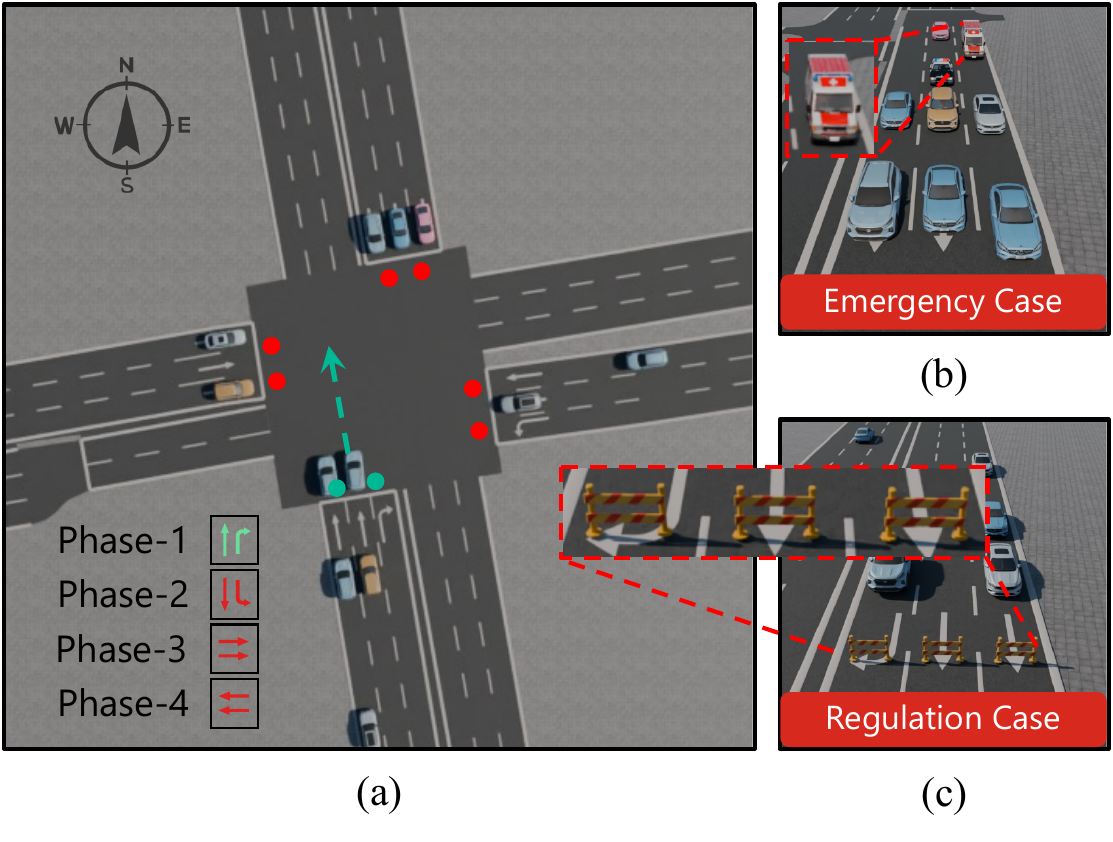}
    \caption{Illustration of the intersection environment and the evaluated open-world scenarios. (a) A standard four-way intersection layout, detailing directional traffic movements and signal phase modeling. (b) An emergency vehicle passage scenario. (c) A temporary traffic regulation scenario.}
    \label{fig:illustration_env}
\end{figure}

% %%%%%%%%%%%%%%
% Preliminaries
% %%%%%%%%%%%%%%
\section{Preliminaries} \label{sec:preliminaries}

In this section, we formalize the TSC problem by introducing the essential traffic terminology and signal modeling. Subsequently, we detail the specific open-world traffic scenarios and the multimodal simulation environment used to evaluate the zero-shot adaptability of our proposed framework.

\subsection{Intersection Topology and Signal Modeling} \label{sec:traffic_terms}
To illustrate the problem formulation, a standard four-way intersection is adopted in this study. As shown in Fig.~\ref{fig:illustration_env}, the layout of an intersection typically consists of multiple incoming approaches and departure legs. Each approach comprises several lanes that direct traffic into specific directional streams, formally referred to as \emph{movements} (e.g., through, left-turn, and right-turn). Specifically, a movement represents the continuous traffic flow from an incoming lane group to a designated departure leg.

To manage these directional streams, the signal control process is structured around \emph{signal phases}. A signal phase represents a combination of non-conflicting movements that are granted the right-of-way simultaneously during a specific control interval. For instance, as illustrated in Fig.~\ref{fig:illustration_env}(a), Phase-1 is currently green, indicating that the corresponding through and left-turn movements are permitted to proceed. Furthermore, to ensure traffic safety and clear the intersection during phase transitions, a fixed yellow clearance interval is inserted between consecutive phases.

\subsection{Open-World Traffic Scenarios} \label{sec:traffic_scenarios}
Real-world TSC tasks extend far beyond merely minimizing vehicle waiting times under normal conditions; they must also navigate various open-world uncertainties. In this study, we consider two representative long-tail traffic scenarios. First, as illustrated in Fig.~\ref{fig:illustration_env}(b), we evaluate the \emph{emergency vehicle passage} scenario. In this safety-critical event, emergency vehicles (e.g., ambulances) approach the intersection and require immediate priority. This disrupts regular traffic flow and demands the TSC system to visually identify the approaching vehicle and promptly preempt the normal signal cycle to grant safe passage. Second, Fig.~\ref{fig:illustration_env}(c) illustrates the \emph{temporary traffic regulations} scenario, which simulates ad-hoc disruptions such as unexpected lane closures or roadworks. These conditions fundamentally alter the spatial dynamics of the intersection.

The common challenge in both scenarios is their heavy reliance on unstructured visual semantics, which predefined numerical states in standard RL fail to capture. To effectively emulate these visual-dependent conditions, we adopt TransSimHub~\cite{wang2025transimhub}, a vision-enabled traffic simulation platform built upon SUMO~\cite{guastella2023traffic}. By leveraging the diverse multi-view images generated by this platform, we establish these open-world uncertainties as our primary testbed. This multimodal environment allows us to comprehensively demonstrate how integrating visual reasoning can achieve zero-shot adaptability without prior retraining. Detailed experimental configurations, including different intersection layouts, are deferred to Section~\ref{experiment_setup}.

\begin{figure*}[!ht]
	\centering
	\includegraphics[width=0.99\linewidth]{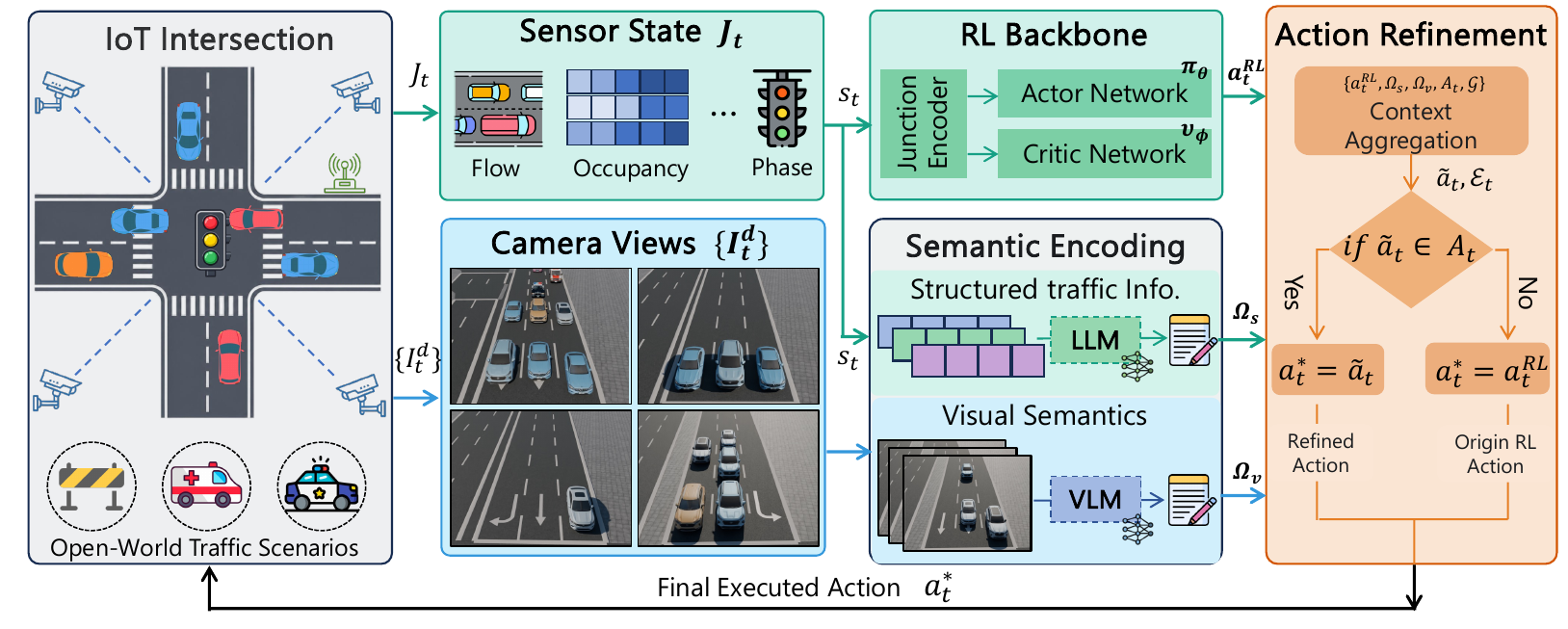}
    \caption{Overall architecture of ReasonLight for zero-shot TSC in IoT-enabled intersections. The RL backbone generates a candidate phase from structured traffic states, while the action refinement block uses camera views and sensor-derived scene descriptions to evaluate whether the RL action should be preserved or refined under available-phase constraints.}
	\label{fig:framework}
\end{figure*}

% %%%%%%%%%%%%
% Methodology
% %%%%%%%%%%%%
\section{Methodology} \label{sec:method}

This section presents ReasonLight, an IoT-oriented multimodal refinement framework for RL-based TSC. We first introduce the overall architecture and inference flow, then describe the Sensor State representation, RL Backbone, Semantic Encoding module, and Action Refinement module.

% Architecture and inference flow
\subsection{ReasonLight Architecture and Inference Flow}

As illustrated in Fig.~\ref{fig:framework}, ReasonLight operates on an IoT intersection equipped with infrastructure sensors and multi-view cameras. At each timestep $t$, the intersection provides a sensor state $\mathbf{J}_t$ and a set of camera views $\{I_t^d\}_{d=1}^{D}$, where $d$ indexes the traffic direction and $D$ is the number of observed directions. The sensor state $\mathbf{J}_t$ records structured traffic measurements such as flow, occupancy, and signal phase status. It is used by the RL backbone to generate a candidate signal phase, and is also combined with the camera views for semantic encoding.

The RL backbone uses a junction encoder, an actor network $\pi_\theta$, and a critic network $\psi_\phi$. Given $s_t$ constructed from recent sensor states, the actor network outputs the RL candidate action $a_t^{RL}$. In parallel, the semantic encoding module produces two complementary contexts: $\Omega_s$ from structured traffic information and $\Omega_v$ from visual semantics.

The action refinement module performs context aggregation over $(a_t^{RL}, \Omega_s, \Omega_v, \mathcal{A}_t, \mathcal{G})$, where $\mathcal{G}$ denotes the predefined traffic rules and $\mathcal{A}_t$ denotes the set of signal phases available under current traffic-safety constraints. It then proposes a refined action $\tilde{a}_t$ and accepts it only if $\tilde{a}_t \in \mathcal{A}_t$; otherwise, the original RL action is retained. The final executed action $a_t^*$ is applied back to the IoT intersection.

% RL Backbone for Action Generation
\subsection{RL Backbone for Candidate Action Generation}
The RL backbone provides the routine-traffic control policy and supplies the candidate action refined by ReasonLight. ReasonLight is policy-agnostic with respect to this backbone: semantic encoding and action refinement do not assume a specific RL algorithm, reward design, or network architecture, but only consume the safety-checked candidate action $a_t^{RL}$. In this study, the backbone learns from the sensor state and remains unchanged when the multimodal modules are added. This subsection defines the state, action, reward, and training objective used by the instantiated backbone.

% --> state
\textbf{State:} 
At each timestep $t$, the intersection state is represented as a structured feature tensor:

\begin{equation}
    \mathbf{J}_t = [\mathbf{m}^1_t, \ldots, \mathbf{m}^{M}_t] \in \mathbb{R}^{M \times D_f},
\end{equation}
where $M$ denotes the total number of movements at the intersection, and each movement $\mathbf{m}^i_t$ is represented by a $D_f$-dimensional feature vector:

\begin{equation}
    \mathbf{m}^i_t = \left[ q^i_t, O^{\text{max},i}_t, O^{\text{mean},i}_t, I^{\text{type}}_i, L_i, I^{\text{green},i}_t, I^{\text{minG},i}_t \right],
\end{equation}
where $q^i_t$ denotes the average flow rate of movement $i$ at time $t$; $O^{\text{max},i}_t$ and $O^{\text{mean},i}_t$ are the maximum and mean occupancy ratios, respectively; $I^{\text{type}}_i \in \{0,1,2\}$ encodes the movement type, corresponding to through, left-turn, and right-turn movements; $L_i$ is the number of lanes; and $I^{\text{green},i}_t$ and $I^{\text{minG},i}_t$ indicate whether movement $i$ is currently under green and whether the minimum-green constraint is satisfied.

To model temporal traffic dynamics, the agent observes a sequence of the past $K$ timesteps:

\begin{equation}
    s_t = [\mathbf{J}_{t-K+1}, \ldots, \mathbf{J}_t] \in \mathbb{R}^{K \times M \times D_f},
\end{equation}
where $K$ denotes the length of the observation window. This temporal stacking mechanism exposes short-term traffic evolution, allowing the policy to capture dependencies in vehicle arrivals, queue propagation, and phase transitions.

\begin{figure*}[!t]
    \centering
    \includegraphics[width=0.99\linewidth]{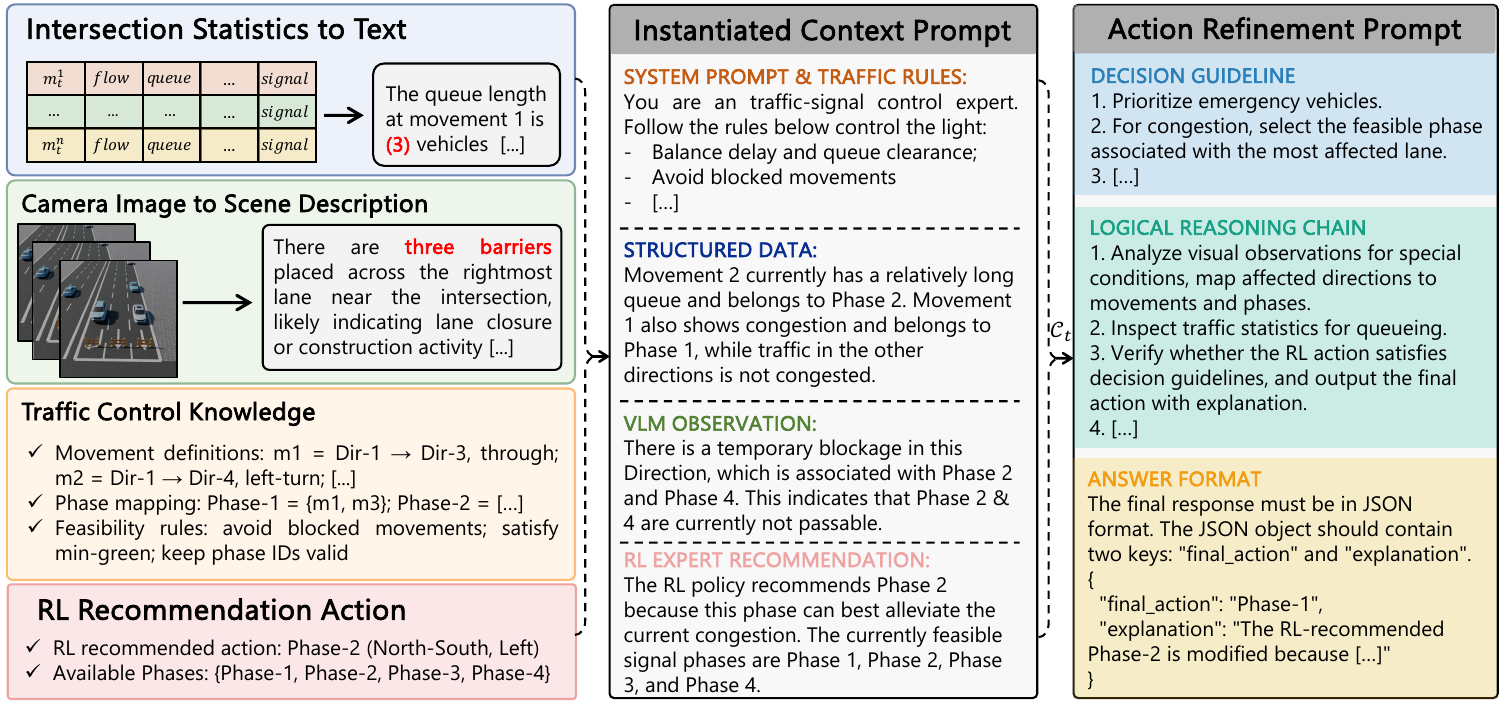}
    \caption{Sequential prompt workflow for action refinement. Multimodal traffic information, traffic rules, the RL-recommended action, and available phases are first assembled into the unified context $\mathcal{C}_t$; a three-component action-refinement prompt then guides $P_{\mathrm{AR}}$ through output-format constraints, logical reasoning, and decision guidelines to produce the final action $a_t^*$.}
    \label{fig:fm-vre}
\end{figure*}

% --> action
\textbf{Action:}
The action space is a predefined discrete phase set:

\begin{equation}
    \mathcal{P} = \{ p_1, p_2, \dots, p_{|\mathcal{P}|} \},
\end{equation}
where $\mathcal{P}$ denotes the set of admissible signal phases and $|\mathcal{P}|$ represents the total number of phases. Each phase $p_k$ corresponds to a predefined group of mutually non-conflicting traffic movements. The cardinality of $\mathcal{P}$ depends on the intersection geometry; for example, $|\mathcal{P}|=4$ for a standard four-leg intersection and $|\mathcal{P}|=3$ for a three-leg (T-shaped) intersection under direction-level phase modeling.

% --> reward
\textbf{Reward:}
To improve traffic efficiency, the reward penalizes average vehicle delay at timestep $t$. Let $\mathcal{V}_t$ denote the set of vehicles present at the intersection, and let $w_t^v$ represent the accumulated waiting time of vehicle $v \in \mathcal{V}_t$. The reward is then defined as:

\begin{equation}
    r_t = - \frac{1}{|\mathcal{V}_t|} 
    \sum_{v \in \mathcal{V}_t} w_t^v.
\end{equation}

Following common RL-based TSC studies, the RL backbone optimizes routine traffic efficiency using average vehicle waiting time as the reward signal. ReasonLight keeps this objective unchanged; zero-shot rare-event adaptation is introduced later through inference-time action refinement rather than retraining or reward redesign.

% --> RL Backbone Inference and Optimization
\textbf{Backbone Inference and Optimization:}
In our implementation, sequential observations $s_t$ are encoded by a junction encoder into a latent representation~\cite{wang2024unitsa}. This representation is shared by the actor network $\pi_\theta$ and the critic network $\psi_\phi$, where the actor outputs signal control actions and the critic estimates the corresponding state values. During inference, the actor maps the observed state to an RL candidate phase:

\begin{equation}
    a_t^{RL} = \pi_\theta(s_t).
\end{equation}

The available phase subset $\mathcal{A}_t \subseteq \mathcal{P}$ is obtained from the intersection phase configuration and current signal-control state, ensuring that actions considered by ReasonLight correspond to admissible phases at the intersection.

In this study, the backbone is optimized using Proximal Policy Optimization (PPO)~\cite{schulman2017proximal}. PPO jointly updates the actor and critic by minimizing the following objective:
\begin{equation}
    \min_{\theta,\phi}
    \; -{\cal L}_{p}(\theta) + \lambda {\cal L}_{v}(\phi),
\end{equation}
where ${\cal L}_{p}$ is the clipped policy objective for the actor, ${\cal L}_{v}$ is the value loss for the critic, and $\lambda$ balances the two terms. The actor term is defined as:
\begin{equation}
{\cal L}_{p}(\theta)=
\mathbb{E}_t \left[
\min \left(
\rho_t(\theta)\hat{A}_t,
\operatorname{clip}(\rho_t(\theta),1-\epsilon,1+\epsilon)\hat{A}_t
\right)
\right],
\end{equation}
where $\rho_t(\theta)=\pi_\theta(a_t^{RL}\mid s_t)/\pi_{\theta_{\mathrm{old}}}(a_t^{RL}\mid s_t)$ is the policy probability ratio between the updated and old actors, and $\epsilon$ is the clipping threshold that limits excessive policy updates. The term $\hat{A}_t$ measures whether the selected action performs better than the critic baseline. In implementation, it is estimated using generalized advantage estimation (GAE):
\begin{equation}
\delta_t = r_t + \gamma \psi_\phi(s_{t+1}) - \psi_\phi(s_t),
\end{equation}
\begin{equation}
\hat{A}_t =
\sum_{\ell=0}^{T-t}
\left(\gamma \lambda_{\mathrm{GAE}}\right)^{\ell}
\delta_{t+\ell},
\end{equation}
where $\lambda_{\mathrm{GAE}}$ controls the bias--variance trade-off in the advantage estimate, and $T$ is the rollout horizon. The corresponding return target for critic training is $\hat{G}_t=\hat{A}_t+\psi_\phi(s_t)$.

The critic term measures the squared error between the value estimate and the return target:
\begin{equation}
{\cal L}_{v}(\phi)=
\mathbb{E}_t \left[
\left(\psi_\phi(s_t)-\hat{G}_t\right)^2
\right],
\end{equation}
where $\hat{G}_t$ is the value target induced by the advantage estimate and $\gamma$ is the discount factor. PPO is used only as the instantiated backbone in our experiments; the ReasonLight refinement interface can be coupled with other RL policies as long as they provide a candidate action $a_t^{RL}$.

% Semantic Encoding and Prompt Construction
\subsection{Semantic Encoding and Prompt Construction}

As illustrated in Fig.~\ref{fig:framework} and Fig.~\ref{fig:fm-vre}, ReasonLight first converts heterogeneous observations into textual descriptions and then assembles them into a unified prompt template for action refinement. This process contains two text-conversion branches and one template-construction step.

% --> image to text
\textbf{Image-to-text conversion:}
At each timestep $t$, camera views are processed by a VLM to produce textual visual semantics rather than raw phase decisions. Let $I_t^d$ denote the image captured from direction $d \in \{1,\dots,D\}$ at timestep $t$. The VLM extracts a direction-level semantic description:
\begin{equation}
v_t^d = f_{\mathrm{VLM}}(I_t^d),
\end{equation}
where $v_t^d$ describes visually observable traffic cues such as emergency vehicle presence, blocked lanes, queue distribution, and abnormal road conditions. The direction-level descriptions are aggregated into a global visual context:
\begin{equation}
\Omega_v = \mathrm{Aggregate}(\{v_t^d\}_{d=1}^{D}),
\end{equation}
which summarizes the visual state of the intersection.

% --> structured data to text
\textbf{Structured-data-to-text conversion:}
Meanwhile, the sensor state $\mathbf{J}_t$ is converted into a compact structured traffic description. This branch uses an LLM to summarize key traffic statistics such as vehicle flow, lane occupancy, current signal status, and minimum-green satisfaction:
\begin{equation}
\Omega_s = f_{\mathrm{LLM}}(\mathbf{J}_t).
\end{equation}

% --> Prompt template construction
\textbf{Prompt template construction:}
The textual visual semantics $\Omega_v$, structured traffic description $\Omega_s$, RL-recommended action $a_t^{RL}$, available phase set $\mathcal{A}_t$, and general traffic rules $\mathcal{G}$ are assembled into a unified prompt context:
\begin{equation}
\mathcal{C}_t =
\mathrm{Template}(\Omega_s, \Omega_v, a_t^{RL}, \mathcal{A}_t, \mathcal{G}),
\end{equation}
where $\mathcal{G}$ contains traffic-control knowledge and feasibility constraints, such as movement definitions, phase-movement mappings, and minimum-green requirements. It specifies what actions are valid under the signal-control setting, while leaving the prioritization among feasible actions to the refinement prompt. The resulting $\mathcal{C}_t$ is passed to the semantic-guided action refinement module.

% Constrained semantic-guided action refinement
\subsection{Semantic-Guided Action Refinement Module}
\label{sec:semantic_action_refinement}

The semantic-guided action refinement module operates over the available phase set $\mathcal{A}_t$ defined above. Given the prompt context $\mathcal{C}_t$ constructed in Fig.~\ref{fig:fm-vre}, we use an LLM-based refinement function $P_{\mathrm{AR}}$ as an implicit semantic evaluator rather than defining a closed-form differentiable utility:
\begin{equation}
\tilde{a}_t, \mathcal{E}_t = P_{\mathrm{AR}}(\mathcal{C}_t),
\end{equation}
where $\tilde{a}_t$ is the candidate refined phase and $\mathcal{E}_t$ is the explanation trace. As summarized in the refinement-prompt part of Fig.~\ref{fig:fm-vre}, $P_{\mathrm{AR}}$ evaluates $\mathcal{C}_t$ together with three fixed prompt components. The \emph{Decision Guidelines} translate traffic-control priorities into preferences for preserving or revising the RL action. The \emph{Logical Reasoning Chain} organizes the comparison among the RL proposal, visual semantics, structured traffic conditions, and feasible phases. The \emph{Output Format Constraint} requires a machine-parseable phase identifier and concise rationale for automatic validity checking.

The refinement function evaluates whether the RL-proposed phase is consistent with structured traffic conditions, visual semantics, and the traffic rules encoded in $\mathcal{G}$, then obtains the final executed action through validity checking:
\begin{equation}
a_t^* =
\begin{cases}
\tilde{a}_t, & \text{if } \tilde{a}_t \in \mathcal{A}_t,\\
a_t^{RL}, & \text{otherwise.}
\end{cases}
\end{equation}

If the generated output cannot be parsed or the candidate phase is not in $\mathcal{A}_t$, the system retries up to $k$ times. If all attempts fail, ReasonLight executes the original RL action $a_t^{RL}$ under the same traffic-safety constraints as the backbone controller. This constrained design reduces model-induced control errors and prevents the action refinement module from issuing an unavailable traffic phase.

\begin{algorithm}[!ht]
\caption{ReasonLight Inference Process}
\label{alg:reasonlight_inference}
\begin{algorithmic}[1]
  \Require $\pi_{\theta}$, $f_{\mathrm{LLM}}$, $f_{\mathrm{VLM}}$, $P_{\mathrm{AR}}$, $\mathcal{G}$, $\mathcal{P}$, maximum attempts $k$
  \Ensure Executed action sequence $\{a_t^*\}$

  \State Initialize the TSC environment
  \For{each timestep $t$}
    \State Observe state $s_t$, views $\{I_t^d\}_{d=1}^{D}$, and phases $\mathcal{A}_t$
    \State Generate RL candidate $a_t^{RL} \leftarrow \pi_{\theta}(s_t)$
    \State Encode traffic state $\Omega_s \leftarrow f_{\mathrm{LLM}}(\mathbf{J}_t)$
    \State Encode visual context $\Omega_v \leftarrow f_{\mathrm{VLM}}(\{I_t^d\}_{d=1}^{D})$
    \State Assemble prompt $\mathcal{C}_t$ from $\Omega_s,\Omega_v,a_t^{RL},\mathcal{A}_t,\mathcal{G}$
    \State Initialize $a_t^* \leftarrow a_t^{RL}$
    \For{$i = 1$ \textbf{to} $k$}
      \State $(\tilde{a}_t,\mathcal{E}_t) \leftarrow P_{\mathrm{AR}}(\mathcal{C}_t)$
      \If{$\tilde{a}_t \in \mathcal{A}_t$}
        \State $a_t^* \leftarrow \tilde{a}_t$
        \State \textbf{break}
      \EndIf
    \EndFor
    \State Execute $a_t^*$
    \State Observe next state $s_{t+1}$
  \EndFor
\end{algorithmic}
\end{algorithm}

\begin{figure*}[!t]
    \centering
    \subfloat[]{
        \includegraphics[width=0.9\textwidth]{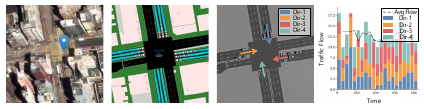}
        \label{fig:env_hk}
    }
    \hfill
    \subfloat[]{
        \includegraphics[width=0.9\textwidth]{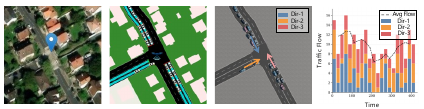}
        \label{fig:env_france}
    }
    \caption{Two real-world-inspired intersection environments used in the experiments: (a) a four-leg intersection in Yau Ma Tei, Hong Kong, and (b) a T-junction in Massy, France.}
    \label{fig:envs}
\end{figure*}

% %%%%%%%%%%%
% Experiment
% %%%%%%%%%%%
\section{Experiment} \label{sec:experiment}

% setup
\subsection{Experimental Setup} \label{experiment_setup}

All experiments are conducted on the TransSimHub~\cite{wang2025transimhub} platform, which integrates the SUMO~\cite{guastella2023traffic} traffic simulator with multi-view camera rendering to provide synchronized structured and visual observations. This setting matches an IoT-enabled intersection deployment, where roadside infrastructure sensors provide quantitative traffic states and cameras provide complementary visual observations. To ensure realistic traffic dynamics, the simulation enforces standard urban traffic regulations, setting the minimum green phase to $10$ seconds and the fixed yellow phase to $3$ seconds. Additionally, a minimum vehicle headway of $2.5$ meters is maintained to simulate safe car-following behaviors.

To evaluate ReasonLight, we select two real-world intersections with distinct topological characteristics. The Yau Ma Tei intersection in Hong Kong, shown in Fig.~\ref{fig:env_hk}, represents a dense urban grid with complex turning restrictions and frequent congestion. The Massy intersection in France, shown in Fig.~\ref{fig:env_france}, is a suburban T-junction featuring wider lanes and simpler traffic patterns. The RL backbone is trained only under routine traffic, while emergency vehicle priority and temporary traffic regulations are evaluated zero-shot without policy retraining, VLM fine-tuning, or scenario-specific prompts. At test time, ReasonLight uses $\Omega_s$ and $\Omega_v$ to refine the RL candidate action $a_t^{RL}$ within the available phase set $\mathcal{A}_t$ and executes the validated action $a_t^*$.

In our implementation, the PPO backbone is trained with a batch size of $128$, a rollout horizon of $T=512$, $10$ optimization epochs per update, a linearly decayed learning rate initialized at $3\times10^{-4}$, a discount factor of $\gamma=0.99$, a generalized-advantage parameter of $\lambda_{\mathrm{GAE}}=0.95$ for estimating $\hat{A}_t$, a clipping threshold of $\epsilon=0.2$, a value-function loss coefficient of $\lambda=0.5$, and an entropy coefficient of $0.01$. For semantic encoding and action refinement, Qwen3-VL~\cite{bai2025qwen3} is deployed as the VLM encoder $f_{\mathrm{VLM}}$, while Qwen3~\cite{yang2025qwen3} is used as the structured encoder $f_{\mathrm{LLM}}$ and the LLM-based refinement function $P_{\mathrm{AR}}$. The resulting average decision latency is approximately $4$ seconds per timestep. ReasonLight is model-agnostic and can integrate other vision-language models with minimal modifications.

\begin{table*}[t]
\caption{Regular-vehicle efficiency under unseen emergency vehicle priority scenarios. $\downarrow$ indicates lower is better; \textsuperscript{\dag}, \textsuperscript{\ddag}, and \textsuperscript{\S} denote first, second, and third within each intersection and metric.}
\label{tab:routine_efficiency}
\centering
\scriptsize
\begin{tabular*}{0.9\textwidth}{@{\extracolsep{\fill}}llcccc@{}}
\toprule
\multicolumn{2}{c}{} & \multicolumn{2}{c}{Yau Ma Tei, Hong Kong} & \multicolumn{2}{c}{Massy, France} \\
\cmidrule(lr){3-4} \cmidrule(lr){5-6}
Category & Model & Avg. Travel Time $\downarrow$ & Avg. Waiting Time $\downarrow$ & Avg. Travel Time $\downarrow$ & Avg. Waiting Time $\downarrow$ \\ \midrule
\multirow{3}{*}{Rule-based}
    & FixTime & $67.63 \pm 4.57$ & $40.00 \pm 2.28$ & $75.84 \pm 4.46$ & $28.19 \pm 1.58$ \\
    & Webster~\cite{webster1958traffic} & $56.26 \pm 3.39$ & $28.62 \pm 1.23$ & $68.92 \pm 2.15$ & $20.89 \pm 0.79$ \\
    & MaxPressure~\cite{varaiya2013max} & $41.36 \pm 2.22$ & $13.33 \pm 0.40$ & $64.82 \pm 4.01$ & $15.25 \pm 0.86$ \\ \midrule
\multirow{4}{*}{RL-based}
    & IntelliLight~\cite{wei2018intellilight} & $69.10 \pm 2.86$ & $13.73 \pm 0.52$ & $58.29 \pm 0.81$ & $10.87 \pm 0.53$ \\
    & UniTSA~\cite{wang2024unitsa} & $\mathbf{38.10 \pm 1.28}$\textsuperscript{\dag} & $\mathbf{10.29 \pm 0.50}$\textsuperscript{\dag} & $\mathbf{57.84 \pm 1.91}$\textsuperscript{\dag} & $10.91 \pm 0.54$\textsuperscript{\ddag} \\
    & CCDA~\cite{wang2024traffic} & $41.60 \pm 2.02$ & $11.23 \pm 0.45$\textsuperscript{\ddag} & $62.83 \pm 2.42$ & $11.19 \pm 0.38$\textsuperscript{\S} \\
    & RL Backbone & $39.54 \pm 1.39$\textsuperscript{\ddag} & $11.51 \pm 0.16$\textsuperscript{\S} & $59.29 \pm 1.41$\textsuperscript{\ddag} & $\mathbf{10.77 \pm 0.18}$\textsuperscript{\dag} \\ \midrule
\multirow{3}{*}{VLM-based}
    & Vanilla-VLM & $62.45 \pm 8.76$ & $17.62 \pm 2.45$ & $68.40 \pm 6.91$ & $18.25 \pm 2.01$ \\
    & VLMLight~\cite{wangvlmlight} & $45.42 \pm 7.18$ & $16.68 \pm 3.10$ & $62.18 \pm 0.58$ & $13.12 \pm 0.13$ \\
    & \textbf{ReasonLight (Ours)} & $41.05 \pm 0.73$\textsuperscript{\S} & $12.73 \pm 0.75$ & $61.89 \pm 0.94$\textsuperscript{\S} & $12.93 \pm 0.37$ \\ \bottomrule
\end{tabular*}
\end{table*}

\begin{table*}[t]
\caption{Emergency vehicle responsiveness under unseen emergency vehicle priority scenarios. All entries report average times for EMVs.}
\label{tab:emv_priority}
\centering
\scriptsize
\begin{tabular*}{0.9\textwidth}{@{\extracolsep{\fill}}llcccc@{}}
\toprule
\multicolumn{2}{c}{} & \multicolumn{2}{c}{Yau Ma Tei, Hong Kong} & \multicolumn{2}{c}{Massy, France} \\
\cmidrule(lr){3-4} \cmidrule(lr){5-6}
Category & Model & EMV Travel Time $\downarrow$ & EMV Waiting Time $\downarrow$ & EMV Travel Time $\downarrow$ & EMV Waiting Time $\downarrow$ \\ \midrule
\multirow{3}{*}{Rule-based}
    & FixTime & $82.67 \pm 3.23$ & $53.17 \pm 2.14$ & $73.60 \pm 4.62$ & $27.80 \pm 1.06$ \\
    & Webster~\cite{webster1958traffic} & $59.67 \pm 4.11$ & $30.83 \pm 1.79$ & $65.20 \pm 3.04$ & $19.80 \pm 0.61$ \\
    & MaxPressure~\cite{varaiya2013max} & $36.17 \pm 2.39$ & $8.83 \pm 0.27$ & $72.40 \pm 2.83$ & $22.40 \pm 0.92$ \\ \midrule
\multirow{4}{*}{RL-based}
    & IntelliLight~\cite{wei2018intellilight} & $65.59 \pm 2.80$ & $8.14 \pm 0.49$ & $57.50 \pm 5.70$ & $7.67 \pm 2.19$ \\
    & UniTSA~\cite{wang2024unitsa} & $33.19 \pm 1.92$ & $5.17 \pm 0.35$\textsuperscript{\S} & $64.90 \pm 2.28$ & $9.81 \pm 0.41$ \\
    & CCDA~\cite{wang2024traffic} & $36.24 \pm 1.45$ & $5.65 \pm 0.31$ & $58.80 \pm 2.40$ & $7.20 \pm 0.49$ \\
    & RL Backbone & $41.03 \pm 0.55$ & $13.22 \pm 1.19$ & $58.08 \pm 2.55$ & $12.27 \pm 0.46$ \\ \midrule
\multirow{3}{*}{VLM-based}
    & Vanilla-VLM & $27.79 \pm 3.79$\textsuperscript{\ddag} & $5.86 \pm 0.69$ & $53.01 \pm 8.34$\textsuperscript{\S} & $3.52 \pm 0.67$\textsuperscript{\S} \\
    & VLMLight~\cite{wangvlmlight} & $32.00 \pm 2.86$\textsuperscript{\S} & $2.57 \pm 0.93$\textsuperscript{\ddag} & $\mathbf{47.32 \pm 6.29}$\textsuperscript{\dag} & $\mathbf{2.04 \pm 2.17}$\textsuperscript{\dag} \\
    & \textbf{ReasonLight (Ours)} & $\mathbf{25.67 \pm 3.30}$\textsuperscript{\dag} & $\mathbf{1.50 \pm 1.81}$\textsuperscript{\dag} & $49.20 \pm 0.92$\textsuperscript{\ddag} & $2.53 \pm 0.50$\textsuperscript{\ddag} \\ \bottomrule
\end{tabular*}
\end{table*}

\begin{table*}[t]
    \centering
    \footnotesize
    \caption{Zero-shot performance under temporary traffic regulations at Yau Ma Tei, Hong Kong.}
    \label{tab:temp_control}
    \begin{tabular*}{0.9\textwidth}{@{\extracolsep{\fill}}lcccc@{}}
        \toprule
        Method & Avg. Travel Time $\downarrow$ & Avg. Waiting Time $\downarrow$ & EMV Travel Time $\downarrow$ & EMV Waiting Time $\downarrow$ \\
        \midrule
        FixTime & $82.60 \pm 0.04$ & $55.49 \pm 0.21$ & $79.67 \pm 4.67$ & $50.67 \pm 4.17$ \\
        RL Backbone & $50.64 \pm 1.22$ & $20.51 \pm 0.87$ & $61.75 \pm 2.47$ & $29.59 \pm 2.00$ \\
        VLMLight & $50.00 \pm 0.08$ & $20.23 \pm 0.11$ & $36.35 \pm 1.91$ & $6.67 \pm 1.65$ \\
        \textbf{ReasonLight (Ours)} & $\mathbf{45.58 \pm 1.56}$ & $\mathbf{16.67 \pm 0.48}$ & $\mathbf{35.59 \pm 1.29}$ & $\mathbf{6.13 \pm 0.06}$ \\
        \bottomrule
    \end{tabular*}
\end{table*}

% Compared methods
\subsection{Compared Methods}

We compare ReasonLight with three categories of baselines:

\textbf{Rule-based methods}: We select three representative strategies: 1) FixTime, a static control policy where we adopt FixTime-30 with a constant phase duration of 30 seconds; 2) Webster~\cite{webster1958traffic}, a classical approach that calculates optimal cycle lengths and phase splits based on traffic demand; and 3) MaxPressure~\cite{varaiya2013max}, a dynamic strategy that greedily selects the phase maximizing the pressure, defined as the difference between upstream and downstream queue lengths.

\textbf{RL-based methods}: We evaluate three existing models and one internal ablation: 1) IntelliLight~\cite{wei2018intellilight}, a deep Q-learning approach operating at 5-second intervals using a balanced replay buffer; 2) UniTSA~\cite{wang2024unitsa}, an RL framework that enhances generalization via junction matrix representations and state augmentation; 3) CCDA~\cite{wang2024traffic}, a centralized-critic and decentralized-actor method adjusting phase durations every 10 seconds; and 4) RL Backbone, an internal ablation of ReasonLight that executes the same actor network $\pi_\theta$ but disables semantic encoding and action refinement.

\textbf{VLM-based methods}: To validate the multimodal design, we compare against: 1) Vanilla-VLM, an ablation baseline where the VLM directly outputs signal phase decisions without the RL module or structured sensor grounding; and 2) VLMLight~\cite{wangvlmlight}, a recent vision-language TSC framework. 

% Metrics
\subsection{Evaluation Metrics}

To evaluate regular-vehicle efficiency, we use Average Travel Time (ATT) and Average Waiting Time (AWT), which quantify how effectively each signal control policy mitigates congestion and reduces delays for non-emergency traffic.

To assess responsiveness to safety-critical traffic, we further report Average Emergency Travel Time (AETT) and Average Emergency Waiting Time (AEWT). These metrics focus on emergency vehicles (EMVs) and directly measure whether a controller can prioritize emergency traffic while preserving feasible signal decisions.

% Performance Comparsion
\subsection{Overall Performance}
As reported in Tables~\ref{tab:routine_efficiency} and~\ref{tab:emv_priority}, together with Table~\ref{tab:temp_control}, we evaluate ReasonLight from three aspects: routine traffic efficiency, emergency vehicle response, and robustness under temporary traffic regulations.

% -> Route Traffic Scenario
\subsubsection{Routine Traffic Efficiency}
Table~\ref{tab:routine_efficiency} evaluates whether ReasonLight preserves regular-vehicle efficiency when emergency vehicle priority scenarios are introduced at test time. Compared with conventional rule-based controllers, ReasonLight consistently achieves lower Avg. Travel Time and Avg. Waiting Time across both intersections. In Hong Kong, it reduces Avg. Waiting Time from $40.00\,\mathrm{s}$ for FixTime and $28.62\,\mathrm{s}$ for Webster to $12.73\,\mathrm{s}$, and slightly improves over MaxPressure ($12.73\,\mathrm{s}$ versus $13.33\,\mathrm{s}$). In France, ReasonLight also outperforms all rule-based methods, reducing Avg. Waiting Time from $15.25\,\mathrm{s}$ for MaxPressure to $12.93\,\mathrm{s}$. These results show that multimodal reasoning does not compromise the basic efficiency requirement of traffic signal control.

Compared with RL-based baselines, ReasonLight shows a modest efficiency trade-off on routine traffic metrics. For example, UniTSA achieves the best routine metrics in Hong Kong, while the RL Backbone obtains the lowest Avg. Waiting Time in France. This is expected because these controllers are optimized primarily for structured-state traffic efficiency, whereas ReasonLight additionally incorporates semantic cues for unseen safety-critical events. Compared with vision-dominant baselines, however, ReasonLight provides more stable routine control: it reduces Avg. Waiting Time from $17.62\,\mathrm{s}$ for Vanilla-VLM and $16.68\,\mathrm{s}$ for VLMLight to $12.73\,\mathrm{s}$ in Hong Kong, and remains comparable to VLMLight in France ($12.93\,\mathrm{s}$ versus $13.12\,\mathrm{s}$). Overall, ReasonLight maintains competitive routine efficiency while enabling the emergency-response capability evaluated next.

\begin{figure*}[!t]
    \centering
    \subfloat[]{
        \includegraphics[width=0.45\textwidth]{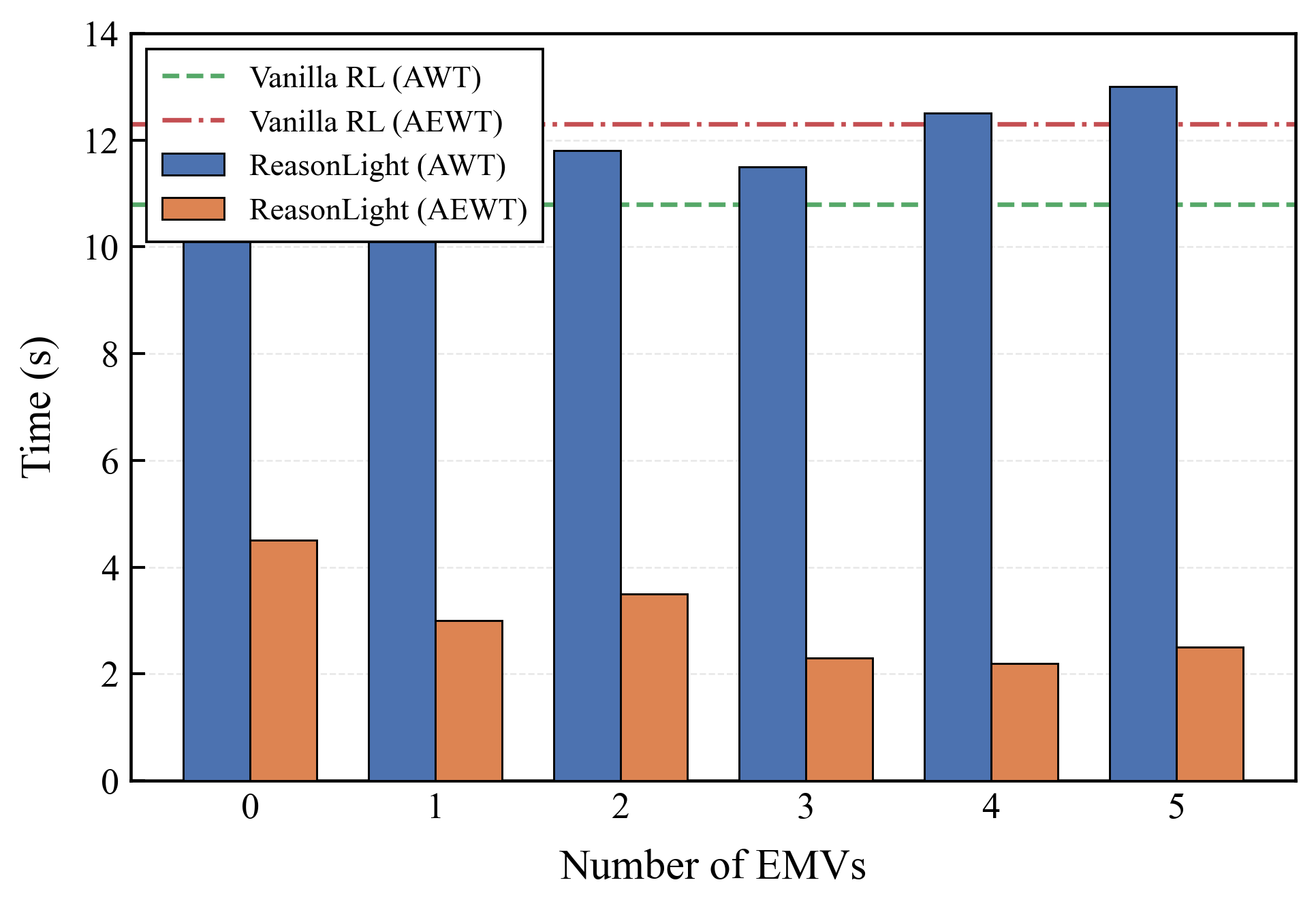}
        \label{fig:emv_efficiency}
    }
    \subfloat[]{
        \includegraphics[width=0.45\textwidth]{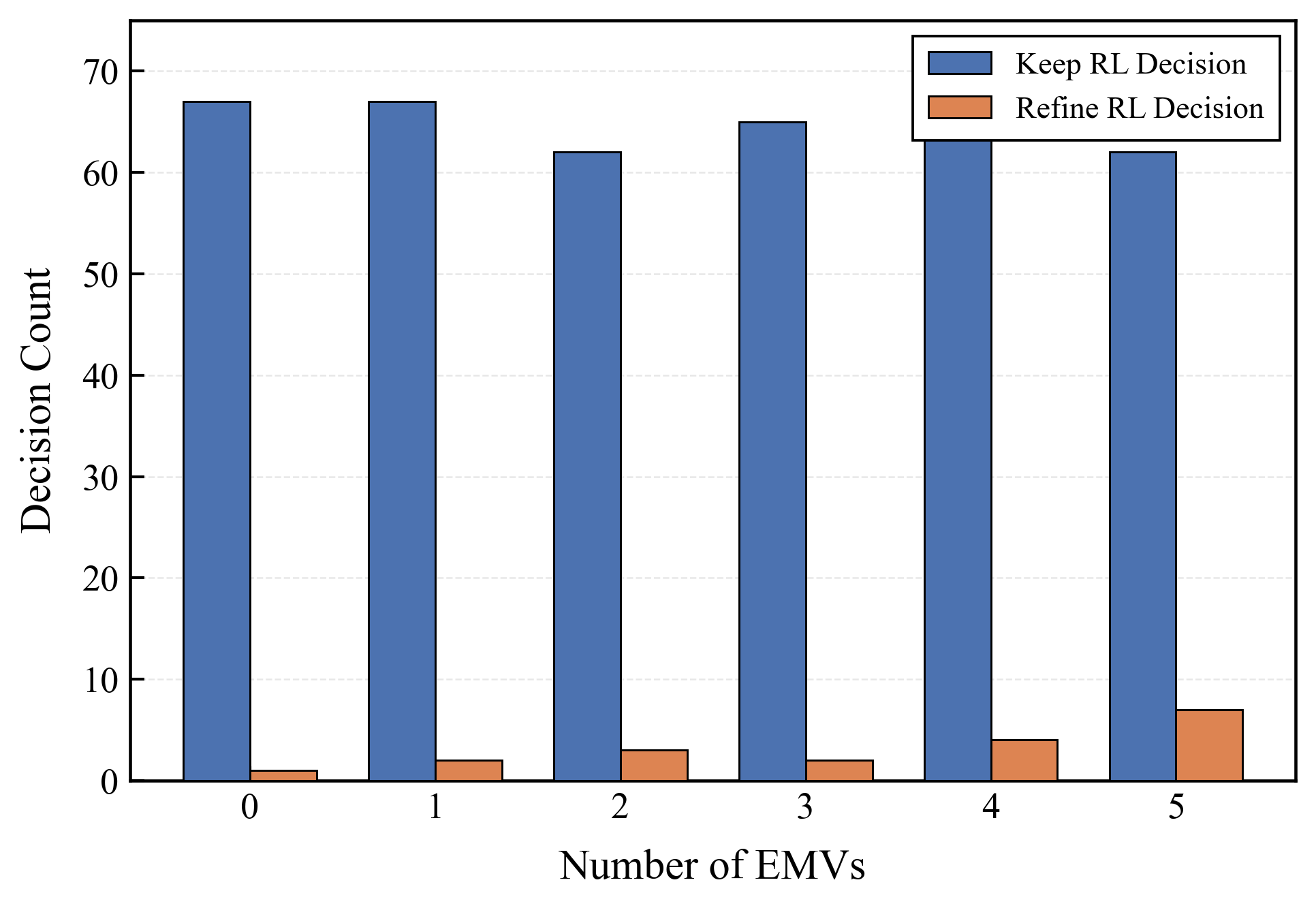}
        \label{fig:emv_refinement}
    }
    \caption{Impact of the number of EMVs on ReasonLight performance: (a) traffic efficiency and (b) number of RL actions preserved by the VLM.}
    \label{fig:diff_env_number}
\end{figure*}

\begin{table*}[t]
    \centering
    \footnotesize
    \caption{Ablation of prompt reasoning components at Yau Ma Tei, Hong Kong.}
    \label{tab:ablation_result}
    \begin{tabular*}{0.9\textwidth}{@{\extracolsep{\fill}}ccc@{\quad}cccc@{}}
      \toprule
      \multicolumn{3}{c}{Modules} & \multicolumn{4}{c}{Metrics} \\
      \cmidrule(lr){1-3} \cmidrule(l){4-7}
      Format & Chain & Guidelines & Avg. Travel Time $\downarrow$ & Avg. Waiting Time $\downarrow$ & EMV Travel Time $\downarrow$ & EMV Waiting Time $\downarrow$ \\
      \midrule
      \cmark & \xmark & \xmark & $48.88$ & $19.92$ & $40.83$ & $11.33$ \\
      \cmark & \xmark & \cmark & $44.90$ & $16.12$ & $33.67$ & $4.67$ \\
      \cmark & \cmark & \cmark & $\mathbf{41.05}$ & $\mathbf{12.73}$ & $\mathbf{25.67}$ & $\mathbf{1.50}$ \\
      \bottomrule
    \end{tabular*}
\end{table*}

% -> Emergency Vehicle Scenario
\subsubsection{Emergency Vehicle Efficiency}
Table~\ref{tab:emv_priority} explains the benefit behind this routine-efficiency trade-off by reporting emergency vehicle responsiveness. Compared with the RL Backbone, ReasonLight substantially reduces EMV Waiting Time from $13.22\,\mathrm{s}$ to $1.50\,\mathrm{s}$ in Hong Kong and from $12.27\,\mathrm{s}$ to $2.53\,\mathrm{s}$ in France, corresponding to improvements of approximately $88.7\%$ and $79.4\%$, respectively. Since the RL policy is not retrained and the structured numerical state does not explicitly encode emergency-vehicle semantics, these gains come from using visual context to refine the RL candidate action under the available phase constraints.

Compared with VLM-based baselines, ReasonLight is strongest in the more complex Hong Kong intersection, where it achieves the lowest EMV Travel Time and EMV Waiting Time. In France, VLMLight obtains lower EMV Travel Time and EMV Waiting Time, indicating that vision-dominant control can be highly responsive in simpler intersection layouts. However, considered together with Table~\ref{tab:routine_efficiency}, ReasonLight provides a stronger balance between routine traffic efficiency and emergency response by grounding visual semantics in structured sensor states and retaining the RL candidate phase as a reliable fallback. This trade-off is important for IoT traffic deployments, where rare-event responsiveness should not destabilize routine traffic operation.

% Temporay Traffic Regulations
\subsubsection{Temporary Regulation Robustness}
To further assess robustness under non-stationary conditions, we introduce a $50\,\mathrm{s}$ period of temporary traffic regulations on the Phase-1 approach at the Yau Ma Tei intersection. This setting represents a blocked-movement scenario: vehicles on the affected approach cannot discharge even when the corresponding signal phase is green. It is fully unseen during RL training and is not described by scenario-specific prompt examples.

As shown in Table~\ref{tab:temp_control}, FixTime and the RL Backbone exhibit clear performance degradation under this disruption, especially in terms of Avg. Waiting Time and EMV Waiting Time. The RL Backbone observes the accumulated queue and waiting time on the affected movement, but it lacks visual information indicating that the road is blocked. It may therefore assign green time to the congested phase under the usual assumption that green service will release queued vehicles, which becomes ineffective in this scenario. VLMLight can observe the road barrier from visual inputs, but it must infer the affected movement, its corresponding phase, and the quantitative queue demand mainly from vision-language reasoning alone; this makes its decisions less stable when movement-phase relationships must be resolved precisely.

In contrast, ReasonLight achieves the strongest overall performance under temporary traffic regulations by linking the observed blockage with structured traffic states and phase definitions. The visual context helps identify that the affected movement cannot be served normally, while the structured queue, waiting-time, and phase information clarifies which signal actions are impacted and which movements remain serviceable. Compared with VLMLight, ReasonLight reduces Avg. Waiting Time from $20.23\,\mathrm{s}$ to $16.67\,\mathrm{s}$ and Avg. Travel Time from $50.00\,\mathrm{s}$ to $45.58\,\mathrm{s}$, corresponding to improvements of approximately $17.6\%$ and $8.8\%$, respectively. It also maintains the lowest EMV Waiting Time at $6.13\,\mathrm{s}$. These results indicate that integrating visual semantics with structured traffic states enables ReasonLight to make more reliable phase decisions under unseen disruptions by balancing scene-level understanding with traffic-control feasibility.

\begin{figure*}[!ht]
	\centerline{\includegraphics[width=0.9\textwidth]{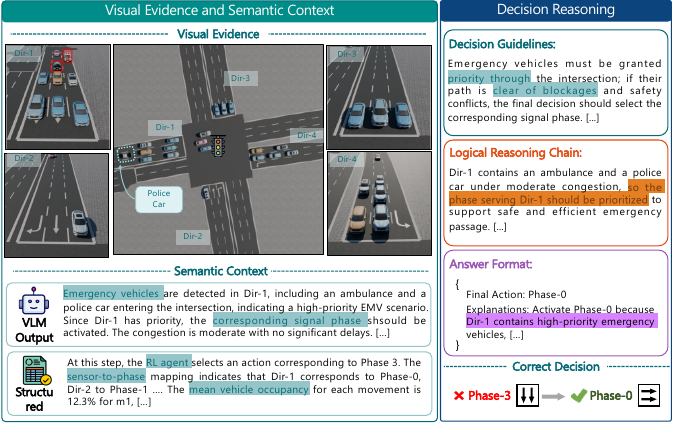}}
	\caption{Case Study 1: ReasonLight Decision Pipeline for Emergency Vehicle Prioritization.}
	\label{fig:case_EMV}
\end{figure*}

% Robustness under Varying Emergency Demand
\subsection{Robustness under Varying Emergency Demand}

Beyond the three scenario-level evaluations above, we further vary the number of EMVs to test whether ReasonLight remains stable as emergency demand increases. Fig.~\ref{fig:diff_env_number} presents the statistical results on how different numbers of EMVs affect system performance. Specifically, Fig.~\ref{fig:emv_efficiency} reports traffic efficiency under varying emergency demand, while Fig.~\ref{fig:emv_refinement} shows the retention of RL-proposed actions after action refinement.

Fig.~\ref{fig:emv_efficiency} shows that increasing emergency demand introduces only a limited impact on the efficiency of regular vehicles. In particular, when no EMV is present in the environment, the AWT of regular vehicles is $11.05\,\mathrm{s}$; when the number of EMVs increases to 5, the AWT rises only to $13.10\,\mathrm{s}$. This indicates that ReasonLight can support stronger emergency priority requirements while maintaining stable routine traffic performance under different traffic pressures.

Fig.~\ref{fig:emv_refinement} further shows the retention of RL-proposed actions after semantic-guided action refinement. As the number of EMVs increases, ReasonLight preserves fewer RL-proposed actions, indicating that the system refines RL actions more frequently to better satisfy the priority requirements of emergency vehicles. Conversely, as the number of EMVs decreases, both the action refinement rate and the AWT gradually decrease. These results indicate that the proposed framework can adaptively adjust its behavior according to observed emergency-demand levels, rather than applying unnecessary semantic intervention under routine traffic conditions.

\begin{figure*}[!ht]
	\centerline{\includegraphics[width=0.9\textwidth]{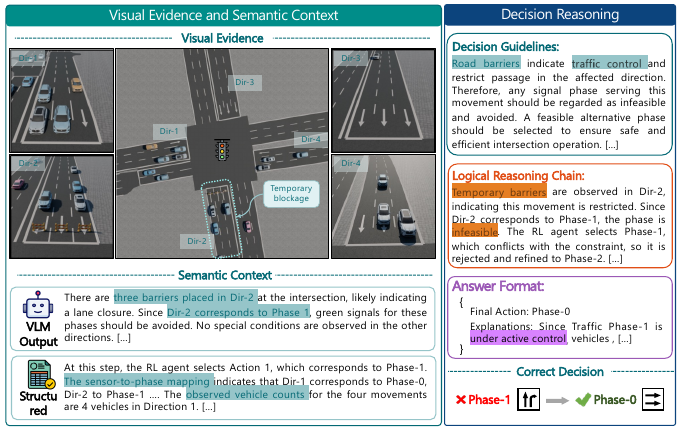}}
	\caption{Case Study 2: ReasonLight Decision Pipeline for Temporary Traffic Regulations.}
	\label{fig:case_traffic_control}
\end{figure*}

% Ablation Analysis
\subsection{Ablation Analysis}

To examine the fixed refinement prompt inside $P_{\mathrm{AR}}$, we conduct ablation studies at the Yau Ma Tei intersection in Hong Kong. All variants share the prompt context $\mathcal{C}_t$, including the traffic rules $\mathcal{G}$, and differ only in the prompt-level controls used by $P_{\mathrm{AR}}$. Since the controller requires a parseable phase identifier, the output format constraint is retained in every variant. We then incrementally add Decision Guidelines and the Logical Reasoning Chain to evaluate their contribution to semantic-guided action refinement. The results are summarized in Table~\ref{tab:ablation_result}.

The format-only variant can produce executable actions, but it lacks explicit priorities for applying the traffic rules to the current decision. As a result, it obtains relatively high Avg. Waiting Time and EMV Waiting Time. Adding Decision Guidelines substantially improves both routine efficiency and emergency response, reducing EMV Waiting Time from $11.33\,\mathrm{s}$ to $4.67\,\mathrm{s}$. This indicates that control-oriented priorities, such as serving feasible emergency-vehicle movements and avoiding blocked movements, help $P_{\mathrm{AR}}$ translate the rules in $\mathcal{G}$ into more traffic-aware refinement decisions.

The full ReasonLight setting further adds the Logical Reasoning Chain, which instructs $P_{\mathrm{AR}}$ to compare the RL-recommended action with visual semantics, structured traffic conditions, and feasible phases before producing the refined action. This yields the best performance, with Avg. Waiting Time and EMV Waiting Time reduced to $12.73\,\mathrm{s}$ and $1.50\,\mathrm{s}$, corresponding to reductions of 36\% and 87\% relative to the format-only setting, respectively. These results support the design in Sec.~\ref{sec:semantic_action_refinement}: prompt-level controls make the LLM-based evaluator more disciplined, auditable, and aligned with traffic-signal constraints.

% Qualitative Case Studies
\subsection{Qualitative Case Studies}

Finally, we examine the reasoning behavior of ReasonLight through two representative case studies: EMV prioritization and temporary traffic regulation, as illustrated in Fig.~\ref{fig:case_EMV} and Fig.~\ref{fig:case_traffic_control}. These two scenarios correspond to safety-critical event handling and dynamic traffic organization changes, respectively. They are used to verify the capability of the proposed framework to integrate multimodal information and make context-aware decisions in complex traffic environments.

In the EMV prioritization case shown in Fig.~\ref{fig:case_EMV}, the visual inputs consist of multi-view observations, denoted as Scene-$i$, where each view corresponds to an incoming direction and its associated signal phase. A bird's-eye view (BEV) is also included to provide global spatial context. ReasonLight identifies the approaching emergency vehicle from visual observations and refines the RL-proposed action when the original phase does not sufficiently prioritize the emergency route. This illustrates how visual semantics compensate for the absence of EMV information in structured sensor states.

In the temporary traffic regulation case shown in Fig.~\ref{fig:case_traffic_control}, lane closures and traffic control measures alter the effective traffic layout, making predefined numerical states insufficient to describe the scene. ReasonLight uses visual cues such as blocked lanes and barrier placement together with structured traffic features, including signal phase, queue length, and waiting time, to evaluate the candidate action. Although the VLM may produce imperfect fine-grained descriptions, such as inaccurate vehicle counts, the final decision remains grounded in structured states and constrained by the available phase set. This case highlights the benefit of combining visual semantics with sensor-based traffic measurements rather than relying on visual reasoning alone.

% %%%%%%%%%%%%
% Conclusion
% %%%%%%%%%%%%
\section{Conclusion} \label{sec:conclusion}

This work presents ReasonLight, an IoT-oriented multimodal TSC framework that integrates RL control, structured traffic sensing, and visual semantic reasoning. By preserving the RL controller as the primary policy and introducing a constrained semantic-guided refinement layer, ReasonLight enhances adaptability to visually observable rare events without requiring retraining or reward redesign. Experiments on two real-world-inspired intersections demonstrate that ReasonLight improves emergency vehicle response and robustness under temporary traffic regulations while maintaining competitive routine traffic performance.

Future work will focus on improving multimodal perception through traffic-domain adaptation of FMs and incorporating V2X communication to further enhance situational awareness. Extending ReasonLight to large-scale multi-intersection coordination and developing continual learning strategies for long-term adaptation are also important directions. Overall, this work highlights the potential of integrating multimodal foundation models with RL for robust, context-aware, and IoT-enabled traffic signal control.

% %%%%%%%%%%
% Reference
% %%%%%%%%%%
\bibliographystyle{IEEEtran}
\bibliography{refs}
\end{document}